\documentclass[letterpaper, 10 pt, conference]{ieeeconf}  

\IEEEoverridecommandlockouts
\usepackage{cite}
\usepackage{amsmath,amsfonts}
\usepackage{algorithm}
\usepackage{algorithmic}
\usepackage{booktabs}
\usepackage{graphicx}
\usepackage{textcomp}
\usepackage{array}
\usepackage{subcaption}
\usepackage{calc}
\usepackage{placeins}
\usepackage[table]{xcolor}
\usepackage{hyperref}
\usepackage{multirow}

\title{\LARGE \bf GBPP: Grasp-Aware Base Placement Prediction for Robots via Two-Stage Learning}

\author{
Jizhuo Chen$^{*}$, 
Diwen Liu$^{*}$, 
Jiaming Wang,
Harold Soh%
\thanks{$^{*}$Equal contribution.}%
\thanks{Emails: \{e0774920, e0905370\}@u.nus.edu, jiaming@comp.nus.edu.sg, harold@comp.nus.edu.sg.}
}

\begin{document}

\maketitle
\thispagestyle{empty}
\pagestyle{empty}

\begin{abstract}
Robots must position their base appropriately to enable successful grasps, yet cluttered and dynamic environments make this a challenging problem. Direct training of learning-based base placement models with large-scale simulation is prohibitively expensive, while purely geometric methods struggle with generality and runtime. To balance these trade-offs, we explore a two-stage approach, which we call \emph{Grasp-Aware Base Placement Prediction (GBPP)}. In the first stage, a lightweight distance--visibility heuristic provides broad coverage by automatically labeling large datasets at negligible cost. In the second stage, a smaller, high-fidelity simulation set refines the model, aligning predictions with real grasp outcomes. Through this process, we empirically study how heuristic bootstrapping and targeted simulation complement each other, and we find that combining them yields practical, data-efficient geometry-aware base placement. In both simulation and real-world evaluations, our system quickly scores hundreds of candidate poses and consistently outperforms geometric baselines, offering lessons on the effective use of low-cost heuristics alongside expensive simulation data.
\end{abstract}

\section{Introduction}
\label{sec:intro}
Where should a mobile robot position itself so that its arm can grasp a target object without collisions or joint-limit violations? In realistic settings such as homes, warehouses, and hospitals, clutter and occlusions render naïve ``drive-until-close'' strategies unreliable, often leaving the robot in unreachable poses and forcing costly re-planning.

Most deployed systems follow modular pipelines: perception localizes the object, navigation drives the base nearby, and a grasp planner attempts to compute a trajectory \cite{liu2024okrobot,chang2023goat}. While convenient, this decomposition neglects the base-placement challenge---navigation ignores arm reach, and grasp planning is constrained by whatever base pose is provided. This mismatch often leads to dead-ends \cite{cohen2024survey}. Geometric extensions such as voxel checks or ray-casting can reduce errors but add high computation costs \cite{saini2024placement}. At the other extreme, task-and-motion planning (TAMP) jointly optimizes base placement, grasp, and trajectory \cite{garrett2021integrated,suwa2024milp}, but runtime and mesh requirements make such solvers difficult to deploy in real-time \cite{migimatsu2020objectcentric}.

\begin{figure*}[tb]
    \centering
    \includegraphics[width=\textwidth]{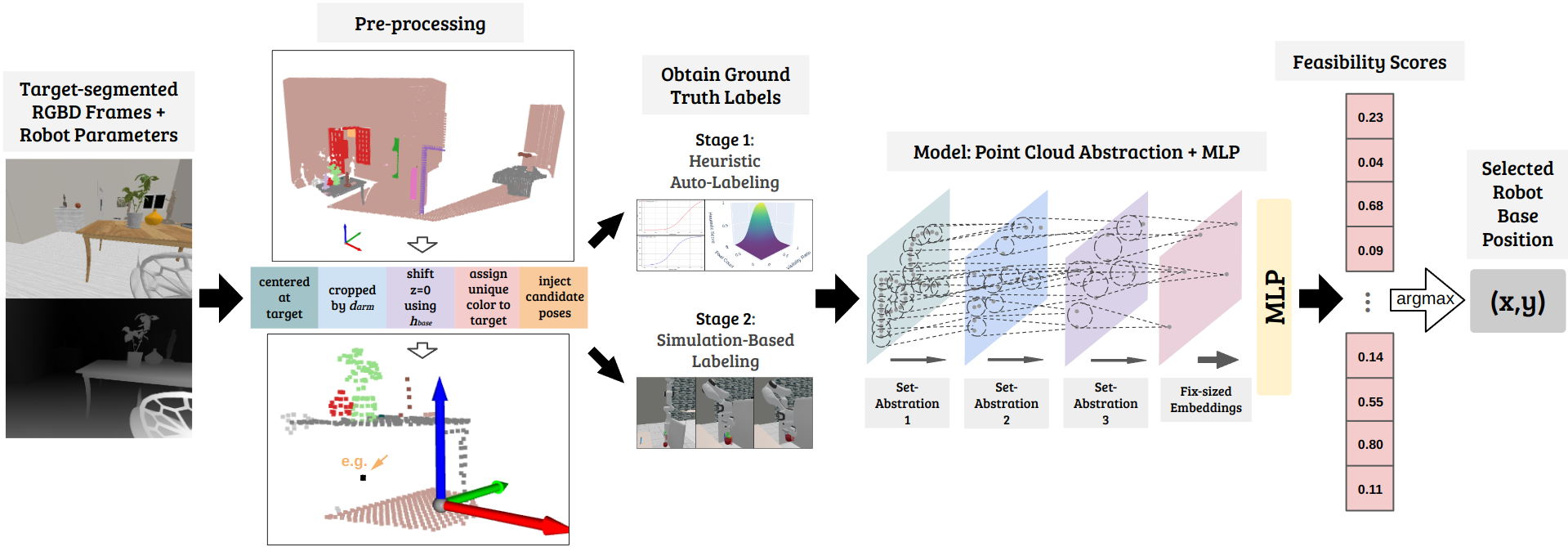}
    \caption{System overview:  
    Input: a target‑segmented RGB-D frame and robot parameters.  
    Model: Three set-abstraction layers as point cloud encoder, with an MLP classifier.  
    Output: feasibility scores for each candidate base pose; the arg‑max is executed.}
    \label{fig:overview}
\end{figure*}

In this work, we study an alternative perspective: casting base placement as a binary classification problem over candidate poses. Given an RGB-D observation, we segment the target, reconstruct its partial point cloud, and combine this with robot parameters. A point-cloud encoder followed by an MLP predicts the feasibility of each candidate, and the robot executes the highest-scoring pose (Fig.~\ref{fig:overview}).  

Training such a model directly with simulation would demand hundreds of thousands of trials, which is computationally prohibitive. Instead, we investigated a two-stage curriculum. A simple distance--visibility rule produced a large pool of inexpensive heuristic labels for initial training, while a smaller, carefully curated set of simulation episodes refined the decision boundary to match true grasp outcomes. Collecting 180k heuristic labels required less than three days on an RTX-A5000, compared to over three weeks for equivalent simulation trials.  

Our results show that this hybrid strategy enables the model to evaluate hundreds of candidate poses in approximately 0.3\,s and generalize from simulation to real-world environments. More importantly, the study highlights what each stage contributes: heuristics provide scale and coverage, while simulation ensures fidelity. Together, they enable practical, data-efficient base placement for mobile manipulators.

\begin{figure}[t] \centering \includegraphics[width=\linewidth]{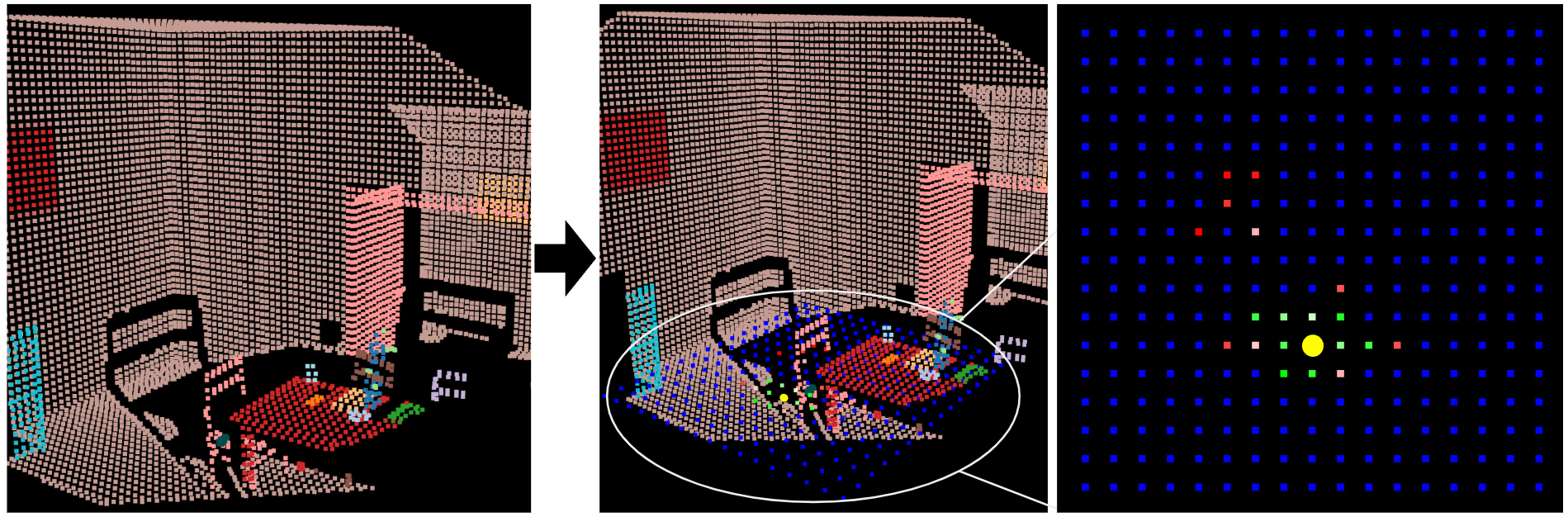} \caption{Given a point cloud and the target object from the robot's observation, GBPP can generate the best robot base position for grasping from a set of candidate positions.} \label{fig:pipeline} \end{figure}

\noindent In summary, this paper should be read less as the presentation of a new algorithm and more as an empirical exploration of design choices and trade-offs in base placement learning. Our contributions are:
\begin{itemize}
    \item We \textbf{empirically examine} the feasibility of combining heuristic bootstrapping with simulation refinement for learning base placement and share insights on the complementary roles of large, inexpensive heuristic data and smaller, high-quality simulation data in shaping decision boundaries.
    \item We \textbf{evaluate trade-offs} between data cost, model performance, and runtime efficiency, providing lessons for future research in mobile manipulation.
\end{itemize}

\section{Related Work} \label{review}

Mobile manipulation research has tackled the challenge of selecting collision‑free base poses, while ensuring targets are within arm reach, through three primary approaches:
(i) modular pipelines that decompose perception, navigation, and grasp planning into separate stages;
(ii) whole‑body task‑and‑motion planning (TAMP) that reasons jointly over base, arm, and grasp; and
(iii) learning‑based scoring that directly predicts the feasibility of each base candidate from sensory data.
Below, we review each approach, highlight their key bottlenecks, and motivate our two-stage curriculum learning solution.

\subsection{Modular Pipelines: Perception–Navigation–Grasp Planning}

Recent service robots, such as OK-Robot~\cite{liu2024okrobot} and GoToAnything~\cite{chang2023goat}, employ modular pipelines where perception modules localize the object, navigation stacks drive the base toward a target location, and grasp planners are invoked only after stopping. This division simplifies software integration and scales well with foundation models. However, the downstream-ignorance inherent in this modular design means navigation often overlooks arm reachability or grasp constraints, resulting in frequent dead-ends that require costly local re-planning~\cite{cohen2024survey}. Attempts to incorporate geometric reasoning - such as ray-casting over voxel maps or precomputed reachability libraries~\cite{makhal2017reuleaux} - are computationally intensive, especially as the density of candidate base poses increases, and can quickly become outdated as environments change~\cite{saini2024placement}. Thus, while modular pipelines enable fast, intuitive navigation, their disregard for manipulation feasibility and the high cost of dense geometry checks limit their effectiveness for robust grasping.

\subsection{Whole‑Body Task‑and‑Motion Planning (TAMP)}

TAMP frameworks abolish the modular boundary by optimizing base pose, grasp pose and joint trajectory in a unified search. Optimization-based or mixed-integer-programming-based solvers guarantee global feasibility in static scenes \cite{garrett2021integrated,suwa2024milp}. However, the price is latency: Suwa et al. \cite{suwa2024milp} report that the GLPK solver requires an average computation time of tens of seconds - up to 75 seconds in their most complex scenarios involving multiple delivery objects and obstacles - for MILP-based fetch-and-carry planning, while Ma et al. observe that sampling-based PDDLStream planners are unable to solve complex, cluttered kitchen benchmarks within the allotted runtime \cite{chen2024pddlstream}. Moreover, although object-centric TAMP frameworks like Migimatsu and Bohg mitigate some issues from perception noise, most TAMP pipelines still assume high‑fidelity meshes or dense signed‑distance fields~\cite{migimatsu2020objectcentric}. Whole‑body optimization offers principled guarantees, but currently delivers second‑to-minute level runtime and presupposes high‑quality geometry that on‑board depth sensors rarely supply.

\subsection{Learning‑based Scoring of Candidate Base Poses}

Learning-based methods bypass explicit optimization by training classifiers to predict grasp feasibility for each candidate base pose directly from perception. Early work relied on hand-crafted distance or visibility heuristics \cite{makhal2017reuleaux}\cite{bircher2016nbv}, while more recent approaches use PointNet++ or graph-based encoders to score gripper poses from large simulation datasets \cite{ni2020pointnetgrasp}\cite{pbssgpd2023}. These methods enable fast inference but often require prohibitively expensive large-scale simulation data for training and typically focus on scoring gripper poses rather than base positions.

In contrast, we adopt a learning-based scoring approach for base placement and address data scarcity through a two-stage learning curriculum: Stage 1 leverages a simple distance–visibility heuristic to auto-label a large, inexpensive training set, while Stage 2 calibrates the model using a smaller, targeted set of high-quality simulation trials. This curriculum, combined with a set-abstraction backbone from PointNet++\cite{qi2017pointnetplusplus} - chosen for its favorable balance between accuracy and throughput compared to heavier transformer variants\cite{zhao2021pointtransformer} - evaluates 600 candidates in approximately 0.3\,s on an RTX‑4090 laptop GPU, enabling dense, real-time evaluation, facilitating sample-efficient, geometry-aware planning in mobile manipulation.


\section{Base Placement Prediction}
\label{sec:method}

\subsection{Task Definition}
\label{ssec:definition}

Given a single RGB-D observation of the environment, the objective is to select a robot base position that enables a collision-free grasp of a segmented target object. We begin by segmenting the target in the image, and then back-projecting to obtain a colored point cloud $\mathcal{P}$ consisting of 3D coordinates and RGB values. Each point $\mathbf{p}_i = (x_i, y_i, z_i, r_i, g_i, b_i)$ is further associated with a binary mask $m_i$, where $m_i = 1$ if the point belongs to the target.

Around the target, we sample a set $\mathcal{B}$ of candidate robot base positions, each denoted $\mathbf{b}_k = (x_k, y_k)$ on the planar workspace. The aim is to choose the optimal base position $\mathbf{b}^{*}$ such that a feasible grasp can be executed:
\[
\mathbf{b}^{*}=\arg\max_{\mathbf{b}_k\in\mathcal{B}} f_{\theta}(\mathbf{b}_k, \mathcal{P}, \mathcal{M}, h_{\mathrm{cam}}, h_{\mathrm{base}}, d_{\mathrm{arm}})
\]
where $f_{\theta}$ is a neural binary classifier that predicts, for each candidate base position $\mathbf{b}_k$, whether the robot can execute a successful grasp given the observed point cloud, mask, camera height ($h_{\mathrm{cam}}$), base height ($h_{\mathrm{base}}$), and the arm's maximum reach ($d_{\mathrm{arm}}$). 

\subsection{Stage 1: Heuristic Auto-Labeling}
\label{ssec:stage1}

To address the challenge of large-scale data collection, we first employ a heuristic-based auto-labeling stage to generate inexpensive training labels for a broad set of base positions. For each candidate base position $\mathbf{b}_k$, we compute its Euclidean distance to the target center ($d_k$), and a visibility score ($\rho_k$), which quantifies the ratio of visible target pixels from $\mathbf{b}_k$ by comparing how much of the target object is visible in the real (cluttered) rendering versus an ideal rendering containing only the object.

The distance score, $\mathrm{distScore}(d)$, is defined by an asymmetric Gaussian centered at a preferred distance $\mu$, with different spreads $\sigma_l, \sigma_r$ for positions nearer or farther than $\mu$:
\[
\mathrm{distScore}(d) =
\begin{cases}
a \exp\!\Bigl(-\frac{(d - \mu)^2}{2\,\sigma_l^2}\Bigr), & \text{if } d < \mu,\\[6pt]
a \exp\!\Bigl(-\frac{(d - \mu)^2}{2\,\sigma_r^2}\Bigr), & \text{if } d \ge \mu,\\[6pt]
0, & \text{if } d \notin [0.14,\,0.92].
\end{cases}
\]

where the range $[0.14, 0.92]$ meters is chosen to encompass the typical reachable workspace of common mobile robot arms. Visibility is mapped via the piecewise mapping:
\[
\mathrm{visScore}(\rho)=
\begin{cases}
0.05 + \dfrac{0.25}{0.04}\rho, & 0 \le \rho < 0.04,\\[4pt]
0.3,                            & 0.04 \le \rho \le 0.8,\\[4pt]
0.3 + \dfrac{0.3}{0.2}(\rho-0.8), & \rho>0.8
\end{cases}
\]
which penalizes low visibility, flattens for moderate views, and rewards near-complete visibility. The combined heuristic score for each candidate is:
\[
H_k = \alpha\,\mathrm{visScore}(\rho_k) + (1-\alpha)\,\mathrm{distScore}(d_k)
\]
where $\alpha = 0.51$ is a hyperparameter tuned to balance the contributions of visibility and distance scores. A global threshold $\tau$ (set to 0.4546 on a validation set to balance FP/FN) is used to produce binary feasibility labels for auto-labeling:
\[
y_k^{\text{heur}} =
\begin{cases}
1, & H_k \ge \tau \\
0, & H_k < \tau
\end{cases}
\]

Each training sample consists of the cropped point cloud (within a radius of $d_{\mathrm{arm}}$ around the target and adjusted so that the $z$-axis is aligned with $h_{\mathrm{base}}$), the binary mask, and the candidate base position $\mathbf{b}_k$. To enable the encoder to infer spatial relationships, each candidate pose is represented as 50 synthetic points at $(x_k, y_k, 0)$ with black color. Points corresponding to the target object are assigned a unique color ($\mathrm{RGB} = [0,77,77]$) that does not occur elsewhere in the point cloud, providing a clear signal for target identification.

We employ 3 Set-Abstraction layers as point cloud encoder, followed by an MLP (layers: 1024$\rightarrow$512$\rightarrow$256$\rightarrow$2 with BatchNorm and Dropout) to process these inputs and output logits $\mathbf{z}_k$ for binary classification. The network is trained on 180k auto-labeled examples using weighted cross-entropy loss, resulting in an initial classifier $f_{\theta}$.

\subsection{Stage 2: Simulation-Based Refinement}
\label{ssec:stage2}

Heuristic labels offer a rough approximation, but may not capture subtle feasibility constraints such as joint limits or complex collisions. Therefore, the second stage refines the classifier using high-quality simulation data.

For each scene in the \texttt{ProcTHOR} \cite{deitke2022procthor} environment, we arrange one target and six clutter YCB \cite{calli2015ycb} objects, then sample a dense grid of candidate base positions (e.g., 25$\times$25). For each candidate $\mathbf{b}_k$, we attempt to plan and execute multiple inverse kinematics (IK) and collision-checked grasp in the \texttt{ManiSkill} \cite{mu2021maniskill} simulator. The simulator returns a ground-truth label $y_k^{\text{sim}}$ indicating grasp feasibility. This process yields approximately 12,000 simulated training examples for the second stage of learning. The classifier $f_{\theta}$ is further trained on these using standard cross-entropy loss.

\subsection{Inference}
\label{ssec:inference}

At test time, given a new scene, the trained classifier evaluates all candidate base positions $\mathbf{b}_k \in \mathcal{B}$, outputs logits $\ell_k$ for the “feasible” class, and selects the highest-scoring, collision-free position for execution:
\[
\mathbf{b}^{*} = \arg\max_{\mathbf{b}_k \in \mathcal{B}} \ell_k
\]

On an RTX-4090 laptop GPU, the network can score 600 candidate poses in approximately 0.3\,s, supporting dense grid evaluation and rapid online planning.


\section{Experiments}
\label{sec:exp}

We evaluate the proposed base-placement predictor in two phases that mirror our two-stage learning curriculum: (i) large‑scale heuristic auto-labeling and (ii) simulation-based refinement. This section details the dataset construction, training protocol, baselines, metrics, and quantitative/qualitative results.

\subsection{Simulation Setup and Data Generation}
\label{ssec:exp_setup}

We use \texttt{ProcTHOR}~\cite{deitke2022procthor} indoor scenes (12,000 layouts) rendered through \texttt{ManiSkill}~\cite{mu2021maniskill} for physics and grasp execution. Each scene contains one target and six clutter YCB~\cite{calli2015ycb} objects. Thirty diverse YCB objects are selected as targets, placed on furniture, with the scene settled for 5,000 physics steps.

For every scene we sample a grid $\mathcal{B}$ of planar base poses around the target. Each candidate $\mathbf{b}_k$ is associated with sampled camera height ($h_{\mathrm{cam}}\!\in\![0.8,1.5]$\,m), base height ($h_{\mathrm{base}}\!\in\![0.5,1.0]$\,m), and arm reach ($d_{\mathrm{arm}}\!\in\![0.3,1.0]$\,m), covering a broad range of real robot configurations. A single RGB-D frame is back‑projected to yield the point cloud $\mathcal{P}$, cropped and aligned as in the methodology.

Stage 1 generates heuristic labels $y_k^{\text{heur}}$ using the distance–visibility scoring rule (Sec.~\ref{ssec:stage1}). Stage 2 uses simulation-derived labels $y_k^{\text{sim}}$ based on grasp feasibility from IK and collision checking in simulation.

The full set ($\approx$252k samples) is partitioned as:
\begin{itemize}
    \item 180,000 heuristic-labeled pairs for Stage~1 training,
    \item 30,000 for validation,
    \item 30,000 for testing,
    \item 12,000 simulation-labelled pairs for Stage~2 and evaluation.
\end{itemize}

\subsection{Training Protocol}
\label{ssec:train_protocol}

In Stage 1, we train three set-abstraction layers with an MLP (1024$\rightarrow$512$\rightarrow$256$\rightarrow$2) with weighted cross-entropy loss on $y_k^{\text{heur}}$. In Stage 2, we continue from Stage~1 and further refine the model on simulation-derived labels $y_k^{\text{sim}}$, using cross-entropy plus weight decay.

\subsection{Baselines}
\label{ssec:baselines}

We use three baselines: (1)
\textbf{Proximity}: Selects the closest candidate base position to the target object, if the base is collision-free; (2) \textbf{Geometric Distance Threshold}: Marks candidates as feasible if they are within the robot’s arm-reach and collision-free. (3) \textbf{Distance Threshold with Object Detector}: Among feasible candidates, select the one with the highest object detection confidence (Detic~\cite{zhou2022detic}).

\subsection{Evaluation Metrics}
\label{ssec:metrics}

Our evaluation metrics comprise:
\begin{itemize}
    \item \textbf{Binary Classification Accuracy}:  Proportion of correct predictions among all candidate poses.
    \item \textbf{Inside / Outside Robot’s Arm-Reach}:   Accuracy reported separately for candidates inside (\(\mathcal{R}_{\text{in}}\)) and outside (\(\mathcal{R}_{\text{out}}\)) the arm-reach threshold.
\item \textbf{Grid Selection Accuracy}: Whether the top-scoring pose $\mathbf{b}^*$ among all candidate poses is truly feasible.
\item \textbf{Nearest-Success Distance (failures only)}: Mean Euclidean distance from an incorrect prediction to the nearest feasible pose.
\end{itemize}

\subsection{Results}
\label{ssec:quant_results}

The Stage 1 model, when both trained and tested on heuristic labels, achieves 87.82\% binary classification accuracy, indicating strong consistency with the auto-labeled data. However, when evaluated against ground-truth simulation labels, performance drops, revealing the limitations of the heuristic. 

Table~\ref{tab:train_regimes} compares binary classification accuracy on simulation labels for all training regimes and baselines. The Stage~1 model achieves $76.75\%$ on simulation labels, reflecting some bias in the heuristic. Our two-stage learning curriculum (heuristic auto-labeling followed by simulation-based refinement) recovers most of this gap, reaching $84.13\%$ - a $7.38\%$ gain over Stage 1 alone. Training exclusively on simulation labels yields $79.11\%$, which is better than Stage~1 alone but below the two-stage model, highlighting the importance of inexpensive data for generalization. For reference, the additional Distance Threshold with Object Detector baseline attains \(68.73\%\) binary accuracy, outperforming plain Proximity baseline \(46.91\%\) but still far below the learned model.

\begin{table}
\centering
\caption{Binary accuracy (\%) on simulation labels for different training regimes and baselines.}
\label{tab:train_regimes}
\begin{tabular}{l c}
\toprule
\textbf{Training Regime / Baseline} & \textbf{Binary Accuracy (\%)} \\
\midrule
Heuristic Labels (Stage 1 Only)     & 76.75 \\
Two-Stage Learning                  & \textbf{84.13} \\
Simulation-only Training (Stage 2)  & 79.11 \\
Proximity                           & 46.91 \\
Geometric Distance Threshold        & 56.91 \\
Distance Threshold + Object Detector& 68.73 \\
\bottomrule
\end{tabular}
\end{table}

Table~\ref{tab:delta_split} breaks simulation accuracy into $\mathcal{R}_{\text{in}}$ (inside the distance threshold) and $\mathcal{R}_{\text{out}}$.  
The Geometric Distance Threshold baseline is trivially perfect outside ($100\%$) because it labels everything far away as infeasible, but it struggles near the object ($56.91\%$).  
Our two‑stage model is markedly better where it matters: $84.13\%$ inside $\mathcal{R}_{\text{in}}$, while essentially matching the baseline outside ($99.17\%$).

\begin{table}
\centering
\caption{Accuracy (\%) with simulation labels, split by distance region ($\mathcal{R}_{\text{in}}$ / $\mathcal{R}_{\text{out}}$).}
\label{tab:delta_split}
\begin{tabular}{l c c}
\toprule
\textbf{Method} & $\text{Acc}_{\mathcal{R}_{\text{in}}}$ (\%) & $\text{Acc}_{\mathcal{R}_{\text{out}}}$ (\%) \\
\midrule
Geometric Distance Threshold & 56.91 & \textbf{100.0} \\
Ours (Two-Stage)             & \textbf{84.13} & 99.17 \\
\bottomrule
\end{tabular}
\end{table}

Finally, Table~\ref{tab:grid_eval} evaluates grid selection accuracy by selecting and evaluating only the top-scoring pose $\mathbf{b}^{*}$, and nearest-success distance when the prediction fails.  
Compared to the proximity baseline, our model more than doubles accuracy ($94.66\%$ vs.\ $46.91\%$) and, when it does err, its chosen pose is much closer to a feasible alternative ($0.24\,$m vs.\ $0.67\,$m).  
This indicates that even incorrect predictions are spatially ``near-misses,'' facilitating quick recovery by local re-planning.

\begin{table}[h]
\centering
\caption{Grid-based selection: accuracy and mean nearest-success distance (for failures).}
\label{tab:grid_eval}
\begin{tabular}{l c c}
\toprule
\textbf{Method} & \textbf{Grid Selection Acc. (\%)} & \textbf{Mean Fail Dist. (m)} \\
\midrule
Proximity Baseline & 46.91 & 0.67 \\
Ours (Two-Stage)    & \textbf{94.66} & \textbf{0.24} \\
\bottomrule
\end{tabular}
\end{table}

Together, these results show that combining heuristic auto-labeling with light simulation-based refinement in a two-stage learning curriculum yields the strongest overall performance, achieving higher binary accuracy than either training stage alone. Our model performs especially well in the challenging near-object region ($\mathcal{R}_{\text{in}}$), where simple baselines fail. Moreover, even incorrect predictions are spatially close to feasible alternatives, supporting quick recovery by local re-planning.

\section{Real‑World Deployment}
\label{sec:real}

We validate that GBPP transfers beyond simulation by deploying it on a Stretch 3 mobile manipulator in cluttered indoor scenes.

\subsection{Experimental Setup}
\label{ssec:real_setup}

\textbf{Robot and sensors}:
we use a Hello Robot Stretch 3 and mount an external Intel RealSense D435i on the mast for ease of calibration and removal. The depth stream is synchronized with RGB to form the single RGB-D snapshot required by our pipeline.

\textbf{Compute and middleware}:
all perception, point‑cloud pre‑processing, candidate scoring, and selection of $\mathbf{b}^{*}$ for final execution run on an RTX‑4090 laptop, which communicates with Stretch over Wi‑Fi via ROS~2. Once $\mathbf{b}^{*}$ is chosen, a navigation goal is issued and the built‑in motion stack handles base motion. The grasp itself is executed using Stretch’s onboard open‑loop grasping routine provided by the Stretch-AI repository ~\cite{stretch_ai2024}, which autonomously adjusts the arm and gripper based on internal heuristics.

\textbf{Pipeline}:
(1) Acquire RGB-D from the D435i and segment the target;  
(2) Back‑project to a colored point cloud, crop by $d_{\mathrm{arm}}$, shift $z{=}0$ using $h_{\mathrm{base}}$, and inject synthetic points for each candidate $\mathbf{b}_k$ (same procedure as described in Sec.~\ref{ssec:exp_setup});  
(3) Run the trained classifier $f_{\theta}$ to obtain logits for all $\mathbf{b}_k$;  
(4) Execute $\mathbf{b}^{*}=\arg\max_k \ell_k$ and perform grasp attempt.  

\textbf{Scene setups:}
We tested three distinct scene layouts: (1) office desk, (2) shelf corner, and (3) living-room table. For each scene, three different targets were selected, and five trials were conducted per target, totaling 15 trials per scene. Success is defined as the robot grasping and lifting the target off the surface at the predicted grasping location.

\begin{figure}[t]
    \centering
    \includegraphics[width=\linewidth]{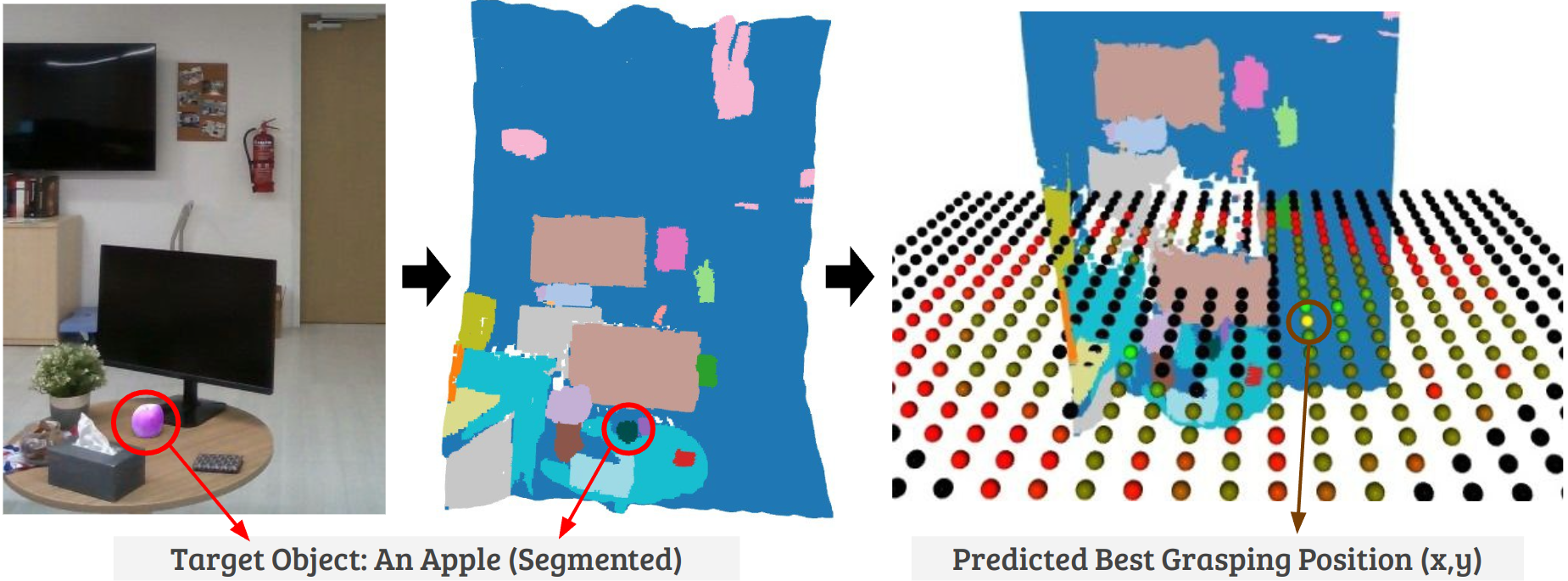}
    \caption{Real‑world deployment: the robot makes RGB-D observation, evaluates candidates and predicts $\mathbf{b}^{*}$.}
    \label{fig:real_pre}
\end{figure}

Figure~\ref{fig:real_pre} shows one full trial from the robot making observations at its initial position to successful pick at the predicted position.

\begin{figure}[t]
    \centering
    \includegraphics[width=\linewidth]{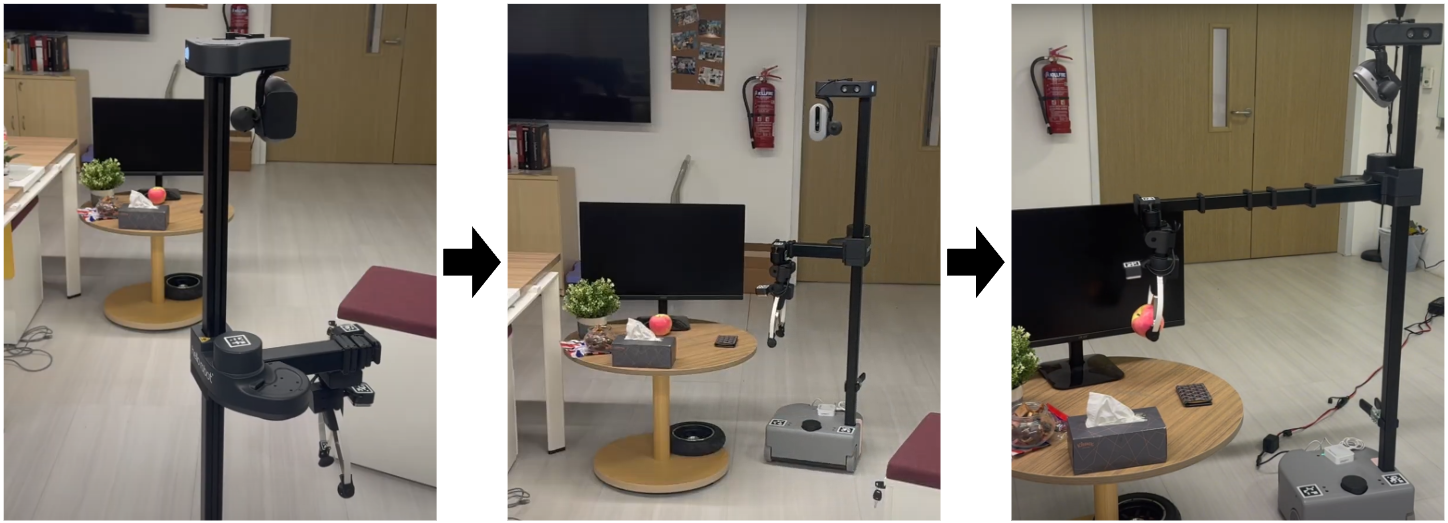}
    \caption{Real‑world deployment: the robot moves to $\mathbf{b}^{*}$ and grasps the target.}
    \label{fig:real_exe}
\end{figure}
\subsection{Results}
\label{ssec:real_results}

Table~\ref{tab:scene_comparison} compares three strategies: 
(1) Proximity Baseline, which selects the closest collision-free candidate base position to the target object based on the point cloud;
(2) an open-loop exploration strategy implemented by the Stretch-AI open-source baseline \cite{stretch_ai2024}, which drives to the nearest unobstructed point in a pre-computed reachability map;
(3) the two-stage GBPP model (ours).

\begin{table}[h]
\centering
\caption{Success rates across three real-world scenes using different methods.}
\label{tab:scene_comparison}
\begin{tabular}{lccc}
\toprule
\textbf{Scene} & \textbf{Proximity} & \textbf{Exploration} & \textbf{GBPP (Ours)} \\
\midrule
Office Desk         & 0.27 & 0.53 & \textbf{0.73} \\
Shelf Corner        & 0.33 & 0.60 & \textbf{0.67} \\
Living-room Table   & 0.40 & 0.73 & \textbf{0.80} \\
\bottomrule
\end{tabular}
\end{table}

The heuristic sometimes parks the robot behind small occluders, while the open-loop exploration strategy remedies obvious collisions but ignores arm kinematics, producing mid-range results. In comparison, GBPP chooses safer stances more consistently, yielding a 73\% overall success rate - $11\%$ over the exploration baseline and $40\%$ over the raw proximity heuristic. End-to-end inference, including ROS overhead, target segmentation, point cloud pre-processing, and location prediction, remains approximately 0.3\,s.

Qualitatively, as seen in Fig.~\ref{fig:real_pre}, the scored grid visualization overlays the real-scene point cloud with a color heatmap ranging from red (low feasibility) to green (high feasibility). The predicted best pose appears as a gold point embedded within a cluster of greens - precisely where visibility and reachability align. The robot navigated to this pose and successfully completed the grasp, demonstrating the model’s ability to localize feasible regions directly from raw RGB-D input.

\subsection{Failure Modes}
\label{ssec:real_failure}

While our real-world experiments show that GBPP can generalize to novel robot platforms and scene setups, one key limitation we identify is its reliance on the quality of input data from consumer-grade RGB-D cameras. In practice, these sensors often produce incomplete or noisy point clouds due to missing depth returns, reflective surfaces, or occlusions caused by clutter. Such imperfections distort the input geometry and can degrade prediction accuracy or cause the model to miss feasible base positions.

\begin{figure}[t]
    \centering
    \includegraphics[width=0.5\linewidth]{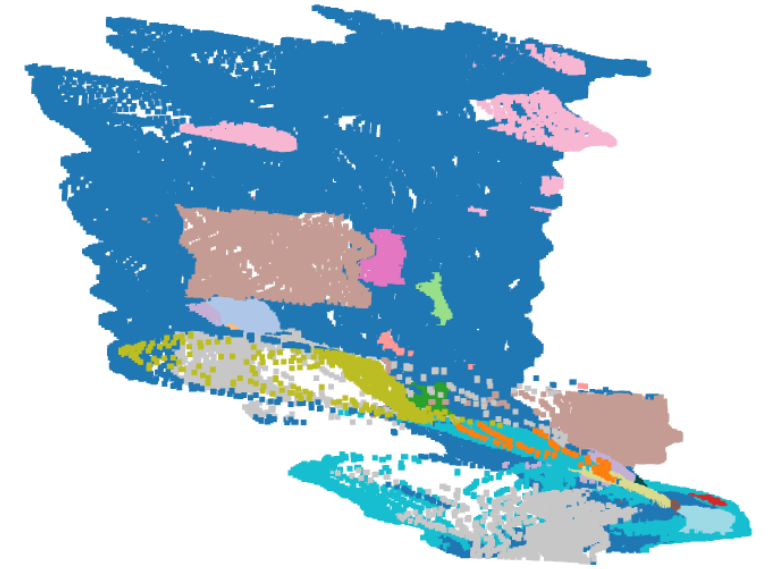}
    \caption{Example of a real-world failure mode: a noisy, incomplete point cloud from a consumer RGB-D camera.}
    \label{fig:real_failure}
\end{figure}

Figure~\ref{fig:real_failure} illustrates a typical failure mode encountered during deployment. A real-world capture from the robot’s RGB-D camera demonstrates significant missing regions and sparsity - especially in areas with reflective or occluded surfaces. This can happen if the depth camera is positioned too far from the scene, or if a lower-cost sensor is used, both of which reduce the quality and density of returned depth data. In this example, the incomplete geometry leads the model to ignore viable base poses near the object, reducing grasp success rates. Overcoming such limitations (for example, via data augmentation, denoising, or multi-view fusion) remains a promising avenue for future work.

\section{Conclusion}
\label{sec:conclusion}

In this work, we studied \emph{Grasp-Aware Base Placement Prediction (GBPP)}, a two-stage curriculum framework for learning geometry-aware base poses in grasping tasks. Rather than relying exclusively on large-scale simulation, which is costly to generate, we combined a simple distance--visibility heuristic for broad, inexpensive auto-labeling with a smaller, high-fidelity simulation set for calibration. This design allowed us to empirically examine how heuristics and simulation complement one another: heuristics provide scale and coverage, while simulation ensures fidelity to true grasp outcomes.  

Our experiments in both simulation and the real world showed that this hybrid approach enables efficient evaluation of candidate base poses and can outperform proximity, geometric, and open-loop baselines across varied environments. More importantly, the study offered insight into the trade-offs between data cost, performance, and generalization, highlighting the value of low-cost heuristics as a foundation for more expensive learning.  

At the same time, our investigation surfaced key limitations. The system remains sensitive to the quality of consumer-grade RGB-D sensors, where missing depth returns, occlusions, and reflective surfaces can distort the point cloud and reduce prediction accuracy. Addressing these perceptual issues---through denoising, sensor fusion, or data augmentation---is an important avenue for future work.  

Overall, our results suggest that pairing heuristic bootstrapping with targeted simulation is a practical path toward scalable and sample-efficient learning for mobile manipulation. Beyond the specific framework, the lessons learned here point to broader opportunities for integrating perception, planning, and learning in complex, cluttered real-world environments.

\newpage{}
\bibliographystyle{IEEEtran}
\bibliography{ref}

@inproceedings{qi2017pointnetplusplus,
  title     = {{PointNet++}: Deep Hierarchical Feature Learning on Point Sets in a Metric Space},
  author    = {Qi, Charles R. and Yi, Li and Su, Hao and Guibas, Leonidas J.},
  booktitle = {Advances in Neural Information Processing Systems (NeurIPS)},
  pages     = {5099--5108},
  year      = {2017}
}

@article{liu2024okrobot,
  title   = {OK-Robot: What Really Matters in Integrating Open-Knowledge Models for Robotics},
  author  = {Liu, Peiqi and Orru, Yaswanth and Paxton, Chris and Shafiullah, Nur Muhammad Mahi and Pinto, Lerrel},
  journal = {arXiv preprint arXiv:2401.12202},
  year    = {2024}
}

@inproceedings{chang2023goat,
  title     = {GOAT: Go To Any Thing --- A Universal Navigation Agent for Open-Vocabulary, Multimodal Goals},
  author    = {Chang, Matthew and Gervet, Theophile and Khanna, Mukul and Yenamandra, Sriram and Shah, Dhruv and Min, So Yeon and Shah, Kavit and Paxton, Chris and Gupta, Saurabh and Batra, Dhruv and Mottaghi, Roozbeh and Malik, Jitendra and Chaplot, Devendra},
  booktitle = {Robotics: Science and Systems (RSS)},
  year      = {2024}
}

@article{cohen2024survey,
  title   = {A Survey of Robotic Language Grounding: Trade-offs between Symbols and Embeddings},
  author  = {Cohen, Vanya and Liu, Jason X. and Mooney, Raymond and Tellex, Stefanie and Watkins, David},
  journal = {arXiv preprint arXiv:2405.13245},
  year    = {2024}
}

@article{saini2024placement,
  title   = {Planning Robot Placement for Object Grasping},
  author  = {Saini, Manish and Jacob, Melvin Paul and Nguyen, Minh and Hochgeschwender, Nico},
  journal = {arXiv preprint arXiv:2405.16692},
  year    = {2024}
}

@article{garrett2021integrated,
  title    = {Integrated Task and Motion Planning},
  author   = {Garrett, Caelan R. and Chitnis, Rohan and Holladay, Rachel and Kim, Beomjoon and Silver, Tom and Kaelbling, Leslie P. and Lozano-P{\'e}rez, Tom{\'a}s},
  journal  = {Annual Review of Control, Robotics, and Autonomous Systems},
  volume   = {4},
  pages    = {265--293},
  year     = {2021},
  publisher= {Annual Reviews}
}

@article{suwa2024milp,
  title   = {Task and Motion Planning Using Mixed Integer Linear Programming for Solving Fetch-and-Carry Tasks by a Mobile Manipulator},
  author  = {Suwa, Sotaro and Takeshita, Keisuke and Yamazaki, Kimitoshi},
  journal = {Advanced Robotics},
  year    = {2024},
  doi     = {10.1080/01691864.2024.2391831}
}

@article{migimatsu2020objectcentric,
  title   = {Object-Centric Task and Motion Planning in Dynamic Environments},
  author  = {Migimatsu, Toki and Bohg, Jeannette},
  journal = {IEEE Robotics and Automation Letters},
  volume  = {5},
  number  = {2},
  pages   = {844--851},
  year    = {2020}
}

@article{makhal2017reuleaux,
  title   = {Reuleaux: Robot Base Placement by Reachability Analysis},
  author  = {Makhal, Abhijit and Goins, Alex K.},
  journal = {arXiv preprint arXiv:1710.01328},
  year    = {2017}
}

@article{chen2024pddlstream,
  title   = {A Task and Motion Planning Framework for Partially Observable Household Manipulation Scenes},
  author  = {Ma, Yuhong and Yuan, Yeqing and Wu, Shaoquan and Yuan, Han},
  journal = {Advanced Intelligent Systems},
  year    = {2025},
  doi     = {10.1002/aisy.202400897},
  note    = {Uses PDDLStream for on-the-fly continuous sampling}
}

@inproceedings{bircher2016nbv,
  title     = {Receding Horizon ``Next-Best-View'' Planner for 3D Exploration},
  author    = {Bircher, Andreas and Kamel, Mina and Alexis, Kostas and Oleynikova, Helen and Siegwart, Roland},
  booktitle = {Proc.\ IEEE International Conference on Robotics and Automation (ICRA)},
  pages     = {1462--1468},
  year      = {2016},
  organization = {IEEE}
}

@inproceedings{ni2020pointnetgrasp,
  title     = {PointNet++ Grasping: Learning an End-to-End Spatial Grasp Generation Algorithm from Sparse Point Clouds},
  author    = {Ni, Peiyuan and Zhang, Wenguang and Zhu, Xiaoxiao and Cao, Qixin},
  booktitle = {Proc.\ IEEE International Conference on Robotics and Automation (ICRA)},
  year      = {2020}
}

@article{pbssgpd2023,
  title   = {Physics-Based Self-Supervised Grasp Pose Detection},
  author  = {Ruiz, Jon A. and Iriondo, Ander and Lazkano, Elena and Ansuategi, Ander and Maurtua, I{\~n}aki},
  journal = {Machines},
  volume  = {13},
  number  = {1},
  pages   = {12},
  year    = {2025},
  doi     = {10.3390/machines13010012}
}

@inproceedings{zhao2021pointtransformer,
  title     = {Point Transformer},
  author    = {Zhao, Hengshuang and Jiang, Li and Jia, Jiaya and Torr, Philip H.\ and Koltun, Vladlen},
  booktitle = {Proc.\ IEEE/CVF International Conference on Computer Vision (ICCV)},
  pages     = {16259--16268},
  year      = {2021}
}

@article{deitke2022procthor,
  title   = {ProcTHOR: Large-Scale Embodied AI Using Procedural Generation},
  author  = {Deitke, Matt and VanderBilt, Eli and Herrasti, {\'A}lvaro and Weihs, Luca and Salvador, Jordi and Ehsani, Kiana and Han, Winson and Kolve, Eric and Farhadi, Ali and Kembhavi, Aniruddha and Mottaghi, Roozbeh},
  journal = {arXiv preprint arXiv:2206.06994},
  year    = {2022}
}

@inproceedings{mu2021maniskill,
  title     = {ManiSkill: Generalizable Manipulation Skill Benchmark with Large-Scale Demonstrations},
  author    = {Mu, Tongzhou and Ling, Zhan and Xiang, Fanbo and Yang, Derek Cathera and Li, Xuanlin and Tao, Stone and Huang, Zhiao and Jia, Zhiwei and Su, Hao},
  booktitle = {Thirty-Fifth Conference on Neural Information Processing Systems Datasets and Benchmarks Track (NeurIPS 2021)},
  year      = {2021}
}

@article{calli2015ycb,
  title   = {The {YCB} Object and Model Set: Towards Common Benchmarks for Manipulation Research},
  author  = {Calli, Berk and Walsman, Aaron and Singh, Arjun and Srinivasa, Siddhartha S. and Abbeel, Pieter and Dollar, Aaron M.},
  journal = {IEEE Robotics \& Automation Magazine},
  volume  = {22},
  number  = {3},
  pages   = {36--52},
  year    = {2015}
}

@inproceedings{zhou2022detic,
  title     = {Detecting Twenty-thousand Classes Using Image-level Supervision},
  author    = {Zhou, Xingyi and Girdhar, Rohit and Joulin, Armand and Kr{\"a}henb{\"u}hl, Philipp and Misra, Ishan},
  booktitle = {European Conference on Computer Vision (ECCV)},
  year      = {2022}
}

@misc{stretch_ai2024,
  title        = {Stretch AI --- an open-source toolkit for embodied intelligence on the Stretch 3 mobile manipulator},
  author       = {{Hello Robot}},
  howpublished = {\url{https://github.com/hello-robot/stretch_ai}},
  year         = {2024},
  note         = {Apache 2.0 / MIT licensed toolkit for grasping, manipulation, navigation, LLM agents}
}

\end{document}